\pdfoutput=1
\documentclass[11pt]{article}

\usepackage[final]{coling}

\usepackage{supertabular}

\usepackage{times}
\usepackage{latexsym}

\usepackage[T1]{fontenc}
\usepackage[utf8]{inputenc}

\usepackage{microtype}

\usepackage{inconsolata}

\usepackage{graphicx}

\title{Lexicography Saves Lives (LSL): Automatically Translating Suicide-Related 
Language}

\author{
  \textbf{Annika Marie Schoene\textsuperscript{1}},
  \textbf{John E. Ortega \textsuperscript{1}},
  \textbf{Rodolfo Joel Zevallos\textsuperscript{2}},
  \textbf{Laura Haaber Ihle \textsuperscript{1}},
\\
  \textsuperscript{1}Northeastern Universty, Institute for Experiential AI,
  \textsuperscript{2}Barcelona Supercomputing Center,
\\
  \small{
    \textbf{Correspondence:} \href{mailto:email@domain}{amschoene@gmail.com}
 }
}

\begin{document}
\maketitle
\begin{abstract}
Recent years have seen a marked increase in research that aims to identify or predict risk, intention or ideation of suicide. The majority of new tasks, datasets, language models and other resources focus on \textit{English} and on suicide in the context of Western culture. However, suicide is global issue and reducing suicide rate by 2030 is one of the key goals of the UN's Sustainable Development Goals\footnote{\url{https://sdgs.un.org/goals/goal3}}. Previous work has used \textit{English} dictionaries related to suicide to translate into different target languages due to lack of other available resources. Naturally, this leads to a variety of ethical tensions (e.g.: \textit{linguistic misrepresentation}), where discourse around suicide is not present in a particular culture or country. In this work, we introduce the \textit{`Lexicography Saves Lives Project'} to address this issue and make \textit{three} distinct contributions. First, we outline ethical consideration and provide overview guidelines to mitigate harm in developing suicide-related resources. Next, we translate an existing dictionary related to suicidal ideation into 200 different languages and conduct human evaluations on a subset of translated dictionaries. Finally, we introduce a public website to make our resources available and enable community participation. 
\end{abstract}

\section{Introduction}

Each year more than 700,000 people die by suicide worldwide \cite{world2021suicide}, where for each suicide there are many more attempts\footnote{\url{https://www.who.int/news-room/fact-sheets/detail/suicide}} and often numbers are underestimated due to  under-reporting or misclassification \citep{snowdon2020undercounting}. There are multiple factors at play that contribute, which include but are not limited to (i) social stigma, (ii) cultural and/or (iii) legal concerns \citep{owidsuicide}. There are a variety of efforts that focus on developing new prevention, screening and risk identification strategies to reduce suicide rates not only from the medical, public health and policy community \citep{morrow2022case,little2016national,denneson2016treatment}, but also from Machine Learning (ML) and Natural Language Processing (NLP) community \citep{kim2023identifying,badal2022natural,mccoy2016improving}. One area of this work focuses on detecting suicidal risk, intent or ideation from social media \citep{coppersmith2015quantifying,du2018extracting,schoene2023example} by utilizing keyword detection approaches based on dictionaries \citep{sinha2019suicidal,ji2020suicidal}. However, the vast majority of this work are developed in \textit{English} and for a western culture, where fewer resources exist in other languages and close to none in low-resource languages. This lack of resources has spurred efforts to automatically translate existing dictionary resources from \textit{English} and \textit{Chinese} to \textit{Korean} \citep{lee2020cross} to predict suicidality. Despite some success of such approaches, there are still a number of challenges that limit the usefulness where automatic translations often don't take cultural context or linguistic differences into account \citep{ortega2023research}. For example, when translating  from \textit{English} to \textit{German} the phrase \textit{`my suicide letter'} is translated into \textit{`Mein Selbstmordbrief'}. Whilst this is a grammatical correct translation, the phrase itself would be rarely if at all used in German and a more commonly used phrase would be \textit{`Mein Abschiedsbrief'}. Similarly, there may be words and phrases in a target language that do not exist in the source language and therefore would be missing in the final translated dictionary. This also raises a set of linguistic and ethical concerns, such as the consequences of designing automatic suicide ideation detection resources/tools and their usefulness in practice. 

Despite these challenges, the potential use of machine translation in the detection of suicide cases is a field that deserves further investigation and careful approaches, as it could offer an additional means of support for suicide prevention on a global scale, provided that the mentioned challenges are appropriately addressed. To grow and improve resource and practices around the development of resources that aid suicide ideation detection in different languages we introduce the \textit{`Lexicography Saves Lives'} and make the following contributions:

\begin{itemize}

    \item First, we describe general ethical issues as well as ethical concerns  related to linguistic misrepresentation that arise with automatic translation in the context of developing resources for suicide ideation detection. Then we provide an overview set of ethical guidelines and mitigation strategies that can aid in developing, designing, and implementing suicide-related resources in the future (section \ref{ethics}). 

    \item Next, we translate a seed dictionary proposed by \citet{o2015detecting} into 200 different languages \citep{team2022no} using the Flores 101 evaluation benchmark \citep{flores101} (section \ref{translate}). We then conduct a human evaluation on a subset of \textit{5} translated dictionaries to better understand the quality. We look at \textit{7} different variables, including \textit{Adequacy}, \textit{Fluency}, \textit{Spelling Errors}, \textit{Culture}, \textit{Context}, \textit{Alternative Translations} and \textit{Contributions in local language}. For \textit{5} out of the \textit{7} variables we quantify our findings to give a score that indicates the quality of the translated dictionary (section \ref{eval}).

    \item Finally, we have created a public website to make our translated and evaluated resources available, provide constant progress updates on what the status of each dictionary is in it's respective language and enable community participation (section \ref{website}).
\end{itemize}

\paragraph{Lexicography Saves Lives Project:} Beyond the scope of this paper, we hope that with this project we raise awareness around the development and deployment of old and new resources (e.g.: datasets, lexicons etc.), algorithms and ultimately tools designed for Mental Health, and especially suicide prevention. We very much see this as a starting point of a wider conversation around how as a community we need to advocate for (i) transparency around harms and benefits of doing such work, (ii) true interdisciplinary research, and (iii) community involvement. 

\section{Related Work}

\paragraph{Detecting suicide-related language} Detection methods for suicidal intent, ideation or risk based on machine learning have evolved significantly over the past decades, and various techniques have been employed to enhance model accuracy. Traditionally, feature engineering has been a crucial component of these methods, where features extracted from text using dictionaries play a pivotal role in training machine learning models \citet{sarsam2021lexicon,birjali2017machine,abboute2014mining,okhapkina2017adaptation,ji2022suicidal}. Our efforts are closely related to previous work that investigated methods and resources for suicide ideation detection. Collecting annotated data for mental health related tasks is notoriously difficult and suicidal ideation is no exception. Work in this area usually relies on self-reports \citep{coppersmith2015quantifying}, heuristics based on the presence of specific keywords \citep{du2018extracting} and phrases or words \citep{burnap2017multi,coppersmith2015quantifying} in lexicons \citet{sawhney2018computational}.

More specifically, lexicons  have typically been developed in collaboration with or by domain experts \cite{gaur2019knowledge}, but more recently computational methods have been used generate new lexicons using social media data \cite{lee2022detecting,lv2015creating}. However, the vast majority of research focuses on developing new datasets, resources (e.g.: language models, lexicons) and tasks only in \textit{English} \citep{du2018extracting,schoene2023example,mishra2019snap,sawhney2021towards,cao2019latent}. Some existing work in languages other than \textit{English} often focus on high-resource languages such as \textit{Spanish} \citep{Valeriano2020,ramirez2020detection}, Arabic \citep{hassib2022aradepsu,benlaaraj2022prediction} or \textit{Chinese} \citep{huang2014detecting,lv2015creating} or and very few investigate low-resource languages (e.g.: \textit{Filipino} or \textit{Taglish} \citep{astoveza2018suicidal}).

Other work \citep{moslem1domain,moslem2023adaptive} focused on the use of adaptive machine translation to create translations for domain-specific text in low-resource languages like \textit{Kinyarwanda}. They showed that generation tools such as GPT-3 \cite{brown2020language, ouyang2022training} and Bloom \cite{davis2023bloom} were inadequate when for domain-specific text in low-resource situations. 

Lastly, \citet{lewis2011crisis} provided a cookbook for MT in crisis situations. Their work could be a good guide to consider when translating texts such as suicide or other life-threatening texts because it outlines steps of actions to take during a crisis. While there is no work that completely covers what to do for translation  purposes in suicidal situations in a low-resource community, we appreciate their cookbook and suggest that more work be done like this for suicidal texts.

\paragraph{Ethical considerations for Mental Health and Suicide} Ethics are another critical concern when considering the implementation of suicide detection systems based on machine translation. People's privacy must be rigorously protected, and any approach in this regard should take into account the ethical implications of online surveillance, the collection of sensitive data, and impact of peoples health. In the past, a few efforts have been made to raise awareness around ethical tensions and issues around online suicide prevention tools, methods and approaches \cite{orr2022ethical}. The majority of existing work in this space has investigated and reviewed existing apps, products or platforms that are already in use \cite{gomes2018ethics,martinengo2019suicide,larsen2015use,jha2023identification,braciszewski2021digital} and much of existing recommendations have been based on practical experience rather than foundational work in bioethics or AI ethics. To the best of our knowledge there has been no effort focused on settings where MT systems are used to automatically translate suicide-related language and has taken into considerations linguistic and cultural factors.

subsection{Linguistic Imbalance and Misrepresentation}
Ethical concerns arise from the imbalance between various languages and language groups or from underlying, but often unarticulated, assumptions about what language is and how it functions.

\paragraph{Imbalances} There is an inherit imbalance between high-resource languages and low-resource languages \cite{ortega2023research}. For example, some languages are far better represented in data sources due to the number of speakers of said language or tech imperialism \cite{kwet2019digital}, and therefore provide a much stronger source for training. This ultimately results in better functioning tools within those language groups. In contrast, low-resource languages are often limited by the size of the data sources, which can result in lower functionality of the finished tool for those language groups. An unintended consequence of this problem is that high-resource language speakers will be better served by the tools developed, while communities who belong to the low-resource language speakers will be under-served, resulting in an obvious distributive injustice, especially if speakers of low-resource languages will still be subjected to the tools, but the tools will be less efficient on such language speakers.

\paragraph{Misrepresentation} poses a risk of low functionality in suicide ideation detection tools in low-resource languages. This can occur when a high-resource language forms the only basis of inquiry, against which all other translations are made. For example, one can imagine a simplistic approach to multilingual suicide ideation detection, where \textit{English} is used as a starting point to identify words related to death and suicide. The list of words is then translated into a variety of low-resource languages, and the results are used to form the basis of the automated suicide ideation detection tools without human intervention. The linguistic misrepresentation will follow from an underlying assumption about what language is and how it functions. Namely that words in language refer to objects or strictly defined concepts in the world and that the meaning of each word follows directly from the object/concept it refers to. If that was the case, it would indeed be possible to simply translate the list from one language to another and obtain a well functioning output. However, a distinction made in philosophy of language may assist in establishing why this understanding is problematic. This distinction is between meaning and reference \cite{putnam1981reason}, where the reference of a word to refer to the object it points to, but the meaning of the word to refer to the intention behind the word \cite{speaks2010theories,frege1892sinn,wittgenstein2010philosophical,russell1905denoting}.

Therefore, if multilingual datasets are developed merely by translating words from a high-resource language to a range of low-resource languages, with no other additional initiatives, the result will be a list of words that reference death and suicide. However, it will not encompass the linguistic meaning of suicide and death in each of the languages to which the translation is made. This becomes evident when one considers how many of the common expressions concerning death or suicide that are perfectly meaningful to language users, but do not contain any reference to death or suicide at all; the \textit{English} phrase \textit{`to kick the bucket'}, has the Danish version \textit{`at stille træskoene” (to put down the wooden clogs)'}, and the Turkish \textit{`nallari dikmek” (to put up the horseshoes)'}. Without a thorough understanding of the fact that a large part of the linguistic meaning we establish when talking about suicide or death are purely metaphorical, allegorical, context dependent, and deeply local to language users, suicide ideation tools will never catch the myriads of ways we meaningfully talk about those topics, and therefore never be as efficient as they could be. 

The ethical effect of this linguistic misrepresentation - a type of representational injustice -is also distributive injustice \cite{zalta1995stanford}, because the tool will have lower precision and functionality in low-resource languages, where attention to how death and suicide are encompassed in language has merely been replaced with a checklist of the words that correspond to those in the high resource language used as the baseline.

\subsubsection{Mitigation strategies} There are various ways to mitigate the aforementioned risks, which include but are not limited to (i) paying close attention to the quality and size of data sources, securing as balanced an approach as possible and (ii) letting each language speak for itself in its respective cultural contexts. It is critical, that local speakers of all languages, but especially low-resource languages are actively included in the translation process, not merely to check if the automated translations make sense, but to add words and phrases from their own language that relate to death and suicide, and to delete those automatically generated translations that are not meaningful in their language. The importance of integrating a strong cultural context becomes clear, when one understands that meaning in language arises through the exact, specific context of each language: To put down the wooden clogs, stems from a time when all Danes were farmers and only had one pair of shoes and putting them down meant one was dead \cite{tangherlini2013danish}. Meaning in language does not merely stem from their reference to objects in the world \cite{wittgenstein2010philosophical}, and for that reason only local language speakers have the authority to define language about death and suicide in their language group. An under-representation of language is an under-representation of the cultural context, in which that language exists. It is evident that even within a particular language group there are linguistic variations and the potential for under-representation of specific sub-groups. For that reason, one should aim for as broad a group of collaborators as possible, within each language group. 

\subsection{Design, Autonomy and Justice}

\paragraph{Design} Both the design and implementation of suicide prevention tools demand a thorough focus on ethics and on the potential implications the technologies in question may have on those who are subjected to them. This requires a theoretically founded and methodological approach to ethics, where foundational work in bioethics \cite{beauchamp2008belmont,national2016policy} and AI ethics \cite{floridi2021ethical,canca2020operationalizing} can provide valuable insights. Even when tools are designed with good intention and the aim to help and protect the most vulnerable, they also have the potential to cause harm or wrongdoing to that same group. This should always be acknowledged by those with the power to develop and implement such tools, and all measures should be taken to ensure that no harm comes to those who are most vulnerable to it. Even if this leads to not implementing the tool is the least potentially harmful path forward. This is especially important in settings, where tech-based initiatives (e.g.: platforms, apps etc.) do not not fall under the scope of binding regulations or organizationally enforced ethical procedures \cite{celedonia2021legal}. In fact, no such guidelines are necessarily developed or implemented. Nor does developing (or implementing) suicide prevention tools demand any professional training in mental health or health care in general. In the worst case, this can result in a setting where expertise is low, ethical concerns are overlooked or missing and the output actively targets the most vulnerable. 

\paragraph{Autonomy} A clear and available introduction to how the tools function should be available to everyone subjected to them, just as it should be possible to inquire about the basis on which they reached specific conclusions. Individuals should be clearly informed about the use of suicide ideation detection tools and should explicitly consent to being subjected to them. To ensure individual autonomy, and due to the potentially severe stigmatization of subjects and the risk of unjust representation in the case of false positives, the utmost care should be given to designing any tools that aim to detect suicide ideation, intent or risk. Any tools developed to prevent suicide require close monitoring of personal data to function, and thus entails a potentially severe privacy violation. This trade off between saving lives and closely monitoring personal data should be made only if the tools are actually effective. Ensuring and documenting efficiency is therefore crucial, in order to justify the potential privacy violation involved.  

\paragraph{Justice} It is vital, to ensure that any resources or tools are subjected to ongoing bias and fairness audits, so as to avoid that the potential burdens of the system (false positives) are not disproportionately distributed on particular groups or individuals, but equally distributed across the population. This is especially important in this context, because of the potentially stigmatizing effect of false positives and the emotional harm they may incur. Furthermore, it is essential to ensure that local regulatory and cultural contexts around suicide are taken into consideration. In 25 countries, not only suicide, but suicide attempts are illegal and for example, in the Bahamas, it is punished with life imprisonment \cite{mishara2016legal}. In such a setting, a suicide prevention tool can become an instrument of power, surveilling citizens for signs of criminal behavior. Therefore, the potential positive effects of suicide prevention tools can outweigh the negative effects and careful measures should be taken prior to development.  

Overall, it is crucial to establish strong ethics practices, clear procedures for complying with them, and processes for documenting that compliance. As the tools in question develop over time, this requires an ongoing involvement with their ethical implications and a thorough integration of ethics into all stages of the development and deployment process, from research, to design, development and deployment. This paper does not offer such an extensive ethics framework, but rather provides an overview guidance on the ethical risks involved in designing, developing, and deploying suicide prevention tools, and to clarify the scope of potential ethical pitfalls, when it comes to developing new resources. 

\section{Automatic translation \label{translate}}
The original lexicon containing words and phrases related to suicidal ideation was proposed by \citet{o2015detecting} and includes the following 50 words and phrases. We provide the full list of phrases/words that express suicidal ideation based on work by \citet{sawhney2018computational} and originally developed by \citet{o2015detecting}:

\begin{itemize}
    \item suicidal, kill myself, my suicide letter, end my life, never wake up, suicide pact, die alone, wanna die, why should I continue living, to take my own life, suicide, can't go on, want to die, be dead, better off without me, better off dead, dont want to be here, go to sleep forever, wanna suicide, take my own life, suicide ideation, not worth living, ready to jump, sleep forever, suicide plan, tired of living, die now, commit suicide, thoughts of suicide, depressed, slit my wrist, cut my wrist, slash my wrist, do not want to be here, want it to be over, want to be dead, nothing to live for, ready to die, not worth living, I wish I were dead, kill me now, hit life, think suicide, wanting to die, suicide times, last day, feel pain point, alternate life, time to go, beautiful suicide, hate life 
\end{itemize}

\subsection{Experiments}
We translate from English into 200 languages originally proposed by the No Language Left Behind (NLLB) \citep{nllb2022} evaluation dataset for low-resource languages. The original language list from NLLB can be found in the Appendix (Section \ref{appendix_A}). The automated machine translation (MT) system used for translation purposes is based on the Flores 101 evaluation benchmark \citep{flores101} which was extended to cover 200 languages\footnote{\url{https://github.com/facebookresearch/flores/blob/main/flores200/README.md}}. We use the Fairseq research toolkit\footnote{\url{https://github.com/facebookresearch/fairseq}} \cite{ott2019fairseq} with the transformer-based \cite{vaswani2017attention} pre-trained language model (PLM) baseline which is a multi-language model that accepts English as the input and is capable of translating to 200 languages. We aim in our experiments to specifically validate suicidal language in one direction, English-->\textit{target language}. For future work, we plan on creating several automated heuristics to automatically validate the translations of new entries and update the current ones where needed.

\section{Multi-lingual Lexicon Evaluation \label{eval}}
We evaluate our automatically translated lexicons by using qualitative and  quantitative measures to gain a deeper insight into the quality and appropriateness of our translations.

\subsection{Human Evaluation \label{human_eval}}
We rely on previous work \citep{ortega2023research,castilho2018evaluating,o2017machine} that describe some of the major facets and caveats of evaluating translations into target languages, including variables that are important to languages that are scarce. For this, we introduce five different variables to evaluate the qualitative validity of each lexicon and two additional \textit{free text} variables for contributions. Here we list and described each variable and how its measured for each words/phrase in a translated lexicon:

\begin{itemize}

    \item \textbf{Adequacy} Similar to \citet{castilho2018evaluating} we ask annotators to evaluate how adequate a translation is by asking \textit{`How much of the source text meaning has been retained in the translated language?'}. The goal is to rate on a Likert scale \cite{jebb2021review} from 1 to 4 how much meaning was retained, where 1 corresponds to \textit{no meaning retained} and 4 \textit{all meaning retained}.

    \item \textbf{Fluency} Similar to \textit{Adequacy} we ask annotators how fluent a translation is by asking \textit{How fluent is the translation?}. As before, we employ a Likert scale from 1 to 4, where 1 is \textit{`no fluency'} and 4 is \textit{`native'}. While `adequacy' would typically measure the degree to which the translation reflects the content of a source sentence, here, ``fluency'' measures how fluent a translation reads without referring to either the source sentence or reference translation \cite{graham2017can}.

    \item \textbf{Spelling Errors} For each translated word/phrase we ask annotators to score one point if a translation contains errors, such as `misspelled words', `missing words', `added words' or `incorrect word order'. 
      
    \item  \textbf{Cultural Acceptability} Given that suicidal language is not universal \cite{kirtley2022translating}, we aim to better facilitate target translations in their cultural context. We propose the following approach: for phrases like ``bite the dust'' that may not translate well into other languages. If the lexicon's target language translation does not match the source language's intent, the participant is asked to select ``no'', otherwise when the source language's lexicon matches the target translation, the participant selects ``yes''. \textit{Question: Does this word/phrase occur in your culture?} 
    
    \item \textbf{Context} After verifying if a word/phrase exists in a cultural context, we also want to verify that it exists in the context of language related to suicide. As previously outlined,  if the word/phrase occurs in this cultural context participants can either select ``yes' or ``no' and have the option to add variation if applicable. \textit{ Question: Does this word/phrase occur in the context of suicidal ideation?}
    
\end{itemize}

\paragraph{Free text variables:}

\begin{itemize}
     \item \textbf{Alternative Translation} In addition to the previous variables, we would also like to take into consideration further feedback. Here we participants can either select `no' or add comments related to each entry. 

    \item \textbf{Contributions in local language} In this section we ask participants to add (i) words related to death, (ii) words related to suicide, (iii) expressions/metaphors related to dying and (iv) expressions/metaphors related to suicide.  
\end{itemize}

\subsection{Metrics \label{quant}}
Translating sensitive content such as language related to suicide ideation requires both high translation quality and appropriateness for the target culture and context \cite{kirtley2022translating}. Quantitative evaluation metrics can complement human assessment to provide a comprehensive analysis of translation adequacy. We use quantitative metrics based on the 5 variables proposed in section \ref{human_eval} to measure appropriateness using human judgments. Each metric follows the same formula, however the meaning of each score may differ. Simply put, we take the total number of entries in the dictionary ($N$) and calculate the arithmetic mean $\overline{x}$ using the sum of submitted evaluations $E$:

\begin{equation}
\overline{x} = \frac{\sum_{E}}{N}
\end{equation}

Furthermore, we calculate each metric per submitted dictionary and then take the average of all submissions per language. For scores related to \textit{Adequacy} and \textit{Fluency} a dictionary can score a maximum value of \textit{4}, meaning all meaning and fluency has been retained in the translation respectively. The best score for \textit{Spelling Errors} is \textit{0}, which shows that no spelling errors were made by the MT system. Next, for \textit{Cultural} and \textit{Contextual acceptability}, the best score is \textit{1} showing that all translations are deemed appropriate. 

\subsection{Pilot Evaluation}
Given the large number of lexicons in a variety of languages, we conduct a round of pilot evaluations of our proposed qualitative measures by inviting native speakers of each language to participate. First, we identify annotators for a select number of languages much as was done in previous work by Facebook \cite{costa2022no,barrault2023seamlessm4t}. Each annotator is given (i) the original dictionary alongside the translation and (ii) a codebook with an example evaluation for reference \footnote{Link to code book: made available upon publication}. At this stage, we specifically ask native speakers in the general public, who have no medical, psychological or behavioral health training. For our first round of evaluations we chose \textit{5} languages, where in Table \ref{tab:qual-eval-pilot} we show the languages and respective scores. Each dictionary was evaluated by at least 2 participants. 

\subsection{Results}
In Table \ref{tab:qual-eval-pilot} we report the scores for each translated dictionary, where we find that for both \textit{Adequacy} and \textit{Fluency} translations are high overall (4 out of 5 lexicons score over 3.0 in both categories) with \textit{Danish} having the highest and \textit{Finish} the lowest score. Furthermore, there are fewer \textit{Spelling Errors} in both \textit{Danish} and \textit{German}, whereas \textit{Galician} has the highest score. This could be due to both languages having being better resource representation and therefore functionality is higher compared to low-resource languages. For \textit{Cultural} and \textit{Contextual Acceptability} scores are lowest in \textit{Finish} indicating that the proposed translations of suicide-related language are not applicable either culturally or contextually. 

\begin{table*}[h!]
\small
\begin{center}
\begin{tabular}{ l | c | c| c| c  | c }
\textbf{Language} & \textbf{Adequacy} &\textbf{Fluency} & \textbf{Spelling Errors} &  \textbf{Culture} & \textbf{Context}  \\ 
\hline
Catalan & 3.68 & 3.6 & 3.7 & 0.76 & 0.8   \\ % John/ Rodolfo 
Danish & 3.74 & 3.6 & 0.02 & 0.98 & 0.86  \\ % Laura
Finish & 2.8 & 2.58 & 0.08 & 0.68 & 0.48   \\
%Spanish & & & & & \\ % John/ Rodolfo 
Galician & 3.48 & 3.41 & 0.29 & 0.96 & 0.96  \\ % John/ Rodolfo 
%Quechua & & & & & \\ % John/ Rodolfo 
German & 3.12 & 2.84 & 0.02 & 0.84 & 0.74 \\ % Annika 
\end{tabular}
\end{center}
\caption{Evaluation of five translated dictionaries using quantative metrics introduced in section \ref{quant}. \label{tab:qual-eval-pilot} }
\end{table*}

\paragraph{Alternative Translations} For each translated lexicon, we also asked participants to provide any alternative translations that may exist in addition to the proposed translation. In Table \ref{alternative} we list a sample of alternative translation for each dictionary, where we find that in some instances more colloquial terms, metaphors or analogies are more appropriate instead of the literal translation provided by the MT system.

\begin{table*}[h!]
\small
\begin{center}
\begin{tabular}{ l | l | l| c}
\textbf{Original word/phrase} & \textbf{Proposed Translation} &\textbf{Alternative Translation} & \textbf{Language} \\ 
\hline
to take my own life & prendre la meva pròpia vida & prendre'm la vida & Catalan \\
 & Otan oman henkeni & otan henkeni  & Finish \\
 & para quitar a miña propia vida & para quitarme a vida & Galician  \\
 & Mein eigenes Leben zu nehmen & mich umbringen & German \\

\hline 
 slit my wrist & T'he tallat la mà & M'he tallat la mà & Catalan \\
 & Skær mit håndled & skærer mit håndled over & Danish \\
& cortoume o pulso & cortar o pulso & Galician \\
& Schneide mir das Handgelenk ab & ritzen & German \\

\hline
go to sleep forever & Gå i seng for evigt & Sove for evigt & Danish \\
 & Vai durmir para sempre & vai adormecer para sempre & Galician \\
 & Schlafe für immer & für immer schlafen & German \\
\end{tabular}
\end{center}
\caption{Examples of alternative translations contributed by participants. \label{alternative}}
\end{table*}

\paragraph{Language Contributions} In addition to alternative translations, we have asked for contributions in local languages that may not be covered in the originally proposed dictionary. In Table \ref{contribution} we list examples in \textit{Danish} and \textit{Finish} as we have been given contributions in those languages. As expected, we find that some terminology that is more relevant to the topic in the target language is not represented in the source dictionary. We hope in future iterations of this work we will be able to solicit more contributions to grow resources and make them available to the wider research community. 

\begin{table*}[h!]
\small
\begin{center}
\begin{tabular}{ l | l | l}
\textbf{Language}  & \textbf{Language Contribution } & \textbf{English Translation} \\ 
\hline
Danish & kradse af  & die \\
Danish & stille træskoene & die \\
Danish & forenes med gud  & unite with the lord \\
Danish & springe i døden  & commit suicide by jumping \\
Danish & Dø for egen hånd  & Suicide (die by your own hand) \\
\hline 
Finish &  itsetuho & self-harm \\
Finish &  itsetuhoisuuden tilalla & about to commit self-harm, suicidal \\
\end{tabular}
\end{center}
\caption{Sample of additional language contributions and their translations submitted by participants \label{contribution} }
\end{table*}

\section{Website for Participation \label{website}}
One of the main aims for this work is to increase community engagement and participation to ensure the responsible development of the proposed lexicons. Therefore we have created a platform \footnote{Made available upon publication: \url{www.dummy.com}} with three main objectives: to provide visual summaries of the progress of the project, to disseminate the dictionaries and other resources generated and finally to collect local feedback and new input from users that help us improve results. The visualization problem abstraction framework defined by \citet{munzner2015visualization}, helps defining data visualization problems by addressing what data is being visualized, why is it visualized (tasks), and how is it visualized (visualization idioms, marks and channels). With this platform we aim to address two main high level analysis tasks (why) as defined by the framework: to \textit{present} the current progress of the project and to \textit{produce} feedback data that helps us improve the project. For the \textit{present} task, the website will include interactive visual summaries (Summarization target action from the framework) that will help users understand the current progress of the project and what main areas have been covered. Moreover, the interactive part of this visualization idioms would allow the user to explore the results and find specific insights that are more relevant for their context (Search action tasks). Furthermore, the platform with also help as a feedback collection mechanism (Produce task), which will allow the users to improve the repository and to adjust the data to the specifics characteristics of each culture and language.

\section{Conclusion, Limitations and Future Work}
In this work, we have introduced the \textit{`Lexicography Saves Lives Project'} and a set of broad ethical considerations and guidelines that look at how this work can be carried forward responsibly. Furthermore, we have automatically translated an existing lexicon containing 50 words and phrases related to suicide into 200 language using Flores 101. Subsequently, we conducted a round of human pilot evaluations on a subset of translated lexicons and introduced a set of metrics to score the quality of a translated lexicon. We described some of our initial findings and outlined our website for participation, which will be launched upon publication of this work. In future iterations of this project we hope to address the following limitations: Firstly, we will develop a comprehensive set of ethics guidelines and assessment strategies to provide ongoing and improved monitoring of this project. Next, we will improve the methodology behind the lexicon quality scores and identify a means of measuring overall quality. As part of this effort, we also plan to compare our current scores against existing and more traditional measures of translation quality in MT systems, such as comparing embedding similarities. Finally, we hope to solicit more evaluations and contributions in local languages.

\bibliography{custom}

\begin{thebibliography}{75}
\providecommand{\natexlab}[1]{#1}

\bibitem[{Abboute et~al.(2014)Abboute, Boudjeriou, Entringer, Az{\'e}, Bringay, and Poncelet}]{abboute2014mining}
Amayas Abboute, Yasser Boudjeriou, Gilles Entringer, J{\'e}r{\^o}me Az{\'e}, Sandra Bringay, and Pascal Poncelet. 2014.
\newblock Mining twitter for suicide prevention.
\newblock In \emph{Natural Language Processing and Information Systems: 19th International Conference on Applications of Natural Language to Information Systems, NLDB 2014, Montpellier, France, June 18-20, 2014. Proceedings 19}, pages 250--253. Springer.

\bibitem[{Astoveza et~al.(2018)Astoveza, Obias, Palcon, Rodriguez, Fabito, and Octaviano}]{astoveza2018suicidal}
Ghelmar Astoveza, Randolph Jay~P Obias, Roi Jed~L Palcon, Ramon~L Rodriguez, Bernie~S Fabito, and Manolito~V Octaviano. 2018.
\newblock Suicidal behavior detection on twitter using neural network.
\newblock In \emph{TENCON 2018-2018 IEEE Region 10 Conference}, pages 0657--0662. IEEE.

\bibitem[{Badal and Depp(2022)}]{badal2022natural}
Varsha~D Badal and Colin~A Depp. 2022.
\newblock Natural language processing of medical records: new understanding of suicide ideation by dementia subtypes: Commentary on “suicidal ideation in dementia: associations with neuropsychiatric symptoms and subtype diagnosis” by naismith et al.
\newblock \emph{International Psychogeriatrics}, 34(4):319--321.

\bibitem[{Barrault et~al.(2023)Barrault, Chung, Meglioli, Dale, Dong, Duquenne, Elsahar, Gong, Heffernan, Hoffman et~al.}]{barrault2023seamlessm4t}
Lo{\"\i}c Barrault, Yu-An Chung, Mariano~Cora Meglioli, David Dale, Ning Dong, Paul-Ambroise Duquenne, Hady Elsahar, Hongyu Gong, Kevin Heffernan, John Hoffman, et~al. 2023.
\newblock Seamlessm4t-massively multilingual \& multimodal machine translation.
\newblock \emph{arXiv preprint arXiv:2308.11596}.

\bibitem[{Beauchamp et~al.(2008)}]{beauchamp2008belmont}
Tom~L Beauchamp et~al. 2008.
\newblock The belmont report.
\newblock \emph{The Oxford textbook of clinical research ethics}, pages 149--155.

\bibitem[{Benlaaraj et~al.(2022)Benlaaraj, El~Jaafari, Ellahyani, and Boutaayamou}]{benlaaraj2022prediction}
Oumaima Benlaaraj, Ilyas El~Jaafari, Ayoub Ellahyani, and Idriss Boutaayamou. 2022.
\newblock Prediction of suicidal ideation in a new arabic annotated dataset.
\newblock In \emph{2022 9th International Conference on Wireless Networks and Mobile Communications (WINCOM)}, pages 1--5. IEEE.

\bibitem[{Birjali et~al.(2017)Birjali, Beni-Hssane, and Erritali}]{birjali2017machine}
Marouane Birjali, Abderrahim Beni-Hssane, and Mohammed Erritali. 2017.
\newblock Machine learning and semantic sentiment analysis based algorithms for suicide sentiment prediction in social networks.
\newblock \emph{Procedia Computer Science}, 113:65--72.

\bibitem[{Braciszewski(2021)}]{braciszewski2021digital}
Jordan~M Braciszewski. 2021.
\newblock Digital technology for suicide prevention.
\newblock \emph{Advances in psychiatry and behavioral health}, 1(1):53--65.

\bibitem[{Brown et~al.(2020)Brown, Mann, Ryder, Subbiah, Kaplan, Dhariwal, Neelakantan, Shyam, Sastry, Askell et~al.}]{brown2020language}
Tom Brown, Benjamin Mann, Nick Ryder, Melanie Subbiah, Jared~D Kaplan, Prafulla Dhariwal, Arvind Neelakantan, Pranav Shyam, Girish Sastry, Amanda Askell, et~al. 2020.
\newblock Language models are few-shot learners.
\newblock \emph{Advances in neural information processing systems}, 33:1877--1901.

\bibitem[{Burnap et~al.(2017)Burnap, Colombo, Amery, Hodorog, and Scourfield}]{burnap2017multi}
Pete Burnap, Gualtiero Colombo, Rosie Amery, Andrei Hodorog, and Jonathan Scourfield. 2017.
\newblock Multi-class machine classification of suicide-related communication on twitter.
\newblock \emph{Online social networks and media}, 2:32--44.

\bibitem[{Canca(2020)}]{canca2020operationalizing}
Cansu Canca. 2020.
\newblock Operationalizing ai ethics principles.
\newblock \emph{Communications of the ACM}, 63(12):18--21.

\bibitem[{Cao et~al.(2019)Cao, Zhang, Feng, Wei, Wang, Li, and He}]{cao2019latent}
Lei Cao, Huijun Zhang, Ling Feng, Zihan Wei, Xin Wang, Ningyun Li, and Xiaohao He. 2019.
\newblock Latent suicide risk detection on microblog via suicide-oriented word embeddings and layered attention.
\newblock \emph{arXiv preprint arXiv:1910.12038}.

\bibitem[{Castilho et~al.(2018)Castilho, Moorkens, Gaspari, Sennrich, Way, and Georgakopoulou}]{castilho2018evaluating}
Sheila Castilho, Joss Moorkens, Federico Gaspari, Rico Sennrich, Andy Way, and Panayota Georgakopoulou. 2018.
\newblock Evaluating mt for massive open online courses: A multifaceted comparison between pbsmt and nmt systems.
\newblock \emph{Machine translation}, 32(3):255--278.

\bibitem[{Celedonia et~al.(2021)Celedonia, Corrales~Compagnucci, Minssen, and Lowery~Wilson}]{celedonia2021legal}
Karen~L Celedonia, Marcelo Corrales~Compagnucci, Timo Minssen, and Michael Lowery~Wilson. 2021.
\newblock Legal, ethical, and wider implications of suicide risk detection systems in social media platforms.
\newblock \emph{Journal of Law and the Biosciences}, 8(1):lsab021.

\bibitem[{Coppersmith et~al.(2015)Coppersmith, Leary, Whyne, and Wood}]{coppersmith2015quantifying}
Glen Coppersmith, Ryan Leary, Eric Whyne, and Tony Wood. 2015.
\newblock Quantifying suicidal ideation via language usage on social media.
\newblock In \emph{Joint statistics meetings proceedings, statistical computing section, JSM}, volume 110.

\bibitem[{Costa-juss{\`a} et~al.(2022)Costa-juss{\`a}, Cross, {\c{C}}elebi, Elbayad, Heafield, Heffernan, Kalbassi, Lam, Licht, Maillard et~al.}]{costa2022no}
Marta~R Costa-juss{\`a}, James Cross, Onur {\c{C}}elebi, Maha Elbayad, Kenneth Heafield, Kevin Heffernan, Elahe Kalbassi, Janice Lam, Daniel Licht, Jean Maillard, et~al. 2022.
\newblock No language left behind: Scaling human-centered machine translation.
\newblock \emph{arXiv preprint arXiv:2207.04672}.

\bibitem[{Dattani et~al.(2023)Dattani, Rodés-Guirao, Ritchie, Roser, and Ortiz-Ospina}]{owidsuicide}
Saloni Dattani, Lucas Rodés-Guirao, Hannah Ritchie, Max Roser, and Esteban Ortiz-Ospina. 2023.
\newblock Suicides.
\newblock \emph{Our World in Data}.
\newblock Https://ourworldindata.org/suicide.

\bibitem[{Davis(2023)}]{davis2023bloom}
Susan~L Davis. 2023.
\newblock Bloom: A multilingual open-source model for addressing ai text biases in natural language processing-a comprehensive review.
\newblock \emph{Advances in AI}, 1(1).

\bibitem[{Denneson et~al.(2016)Denneson, Williams, Kaplan, McFarland, and Dobscha}]{denneson2016treatment}
Lauren~M Denneson, Holly~B Williams, Mark~S Kaplan, Bentson~H McFarland, and Steven~K Dobscha. 2016.
\newblock Treatment of veterans with mental health symptoms in va primary care prior to suicide.
\newblock \emph{General Hospital Psychiatry}, 38:65--70.

\bibitem[{Du et~al.(2018)Du, Zhang, Luo, Jia, Wei, Tao, and Xu}]{du2018extracting}
Jingcheng Du, Yaoyun Zhang, Jianhong Luo, Yuxi Jia, Qiang Wei, Cui Tao, and Hua Xu. 2018.
\newblock Extracting psychiatric stressors for suicide from social media using deep learning.
\newblock \emph{BMC medical informatics and decision making}, 18(2):77--87.

\bibitem[{Floridi et~al.(2021)Floridi, Cowls, Beltrametti, Chatila, Chazerand, Dignum, Luetge, Madelin, Pagallo, Rossi et~al.}]{floridi2021ethical}
Luciano Floridi, Josh Cowls, Monica Beltrametti, Raja Chatila, Patrice Chazerand, Virginia Dignum, Christoph Luetge, Robert Madelin, Ugo Pagallo, Francesca Rossi, et~al. 2021.
\newblock An ethical framework for a good ai society: Opportunities, risks, principles, and recommendations.
\newblock \emph{Ethics, governance, and policies in artificial intelligence}, pages 19--39.

\bibitem[{Frege(1892)}]{frege1892sinn}
Gottlob Frege. 1892.
\newblock {\"U}ber sinn und bedeutung.
\newblock \emph{Zeitschrift f{\"u}r Philosophie und philosophische Kritik}, 100:25--50.

\bibitem[{Gaur et~al.(2019)Gaur, Alambo, Sain, Kursuncu, Thirunarayan, Kavuluru, Sheth, Welton, and Pathak}]{gaur2019knowledge}
Manas Gaur, Amanuel Alambo, Joy~Prakash Sain, Ugur Kursuncu, Krishnaprasad Thirunarayan, Ramakanth Kavuluru, Amit Sheth, Randy Welton, and Jyotishman Pathak. 2019.
\newblock Knowledge-aware assessment of severity of suicide risk for early intervention.
\newblock In \emph{The world wide web conference}, pages 514--525.

\bibitem[{Gomes~de Andrade et~al.(2018)Gomes~de Andrade, Pawson, Muriello, Donahue, and Guadagno}]{gomes2018ethics}
Norberto~Nuno Gomes~de Andrade, Dave Pawson, Dan Muriello, Lizzy Donahue, and Jennifer Guadagno. 2018.
\newblock Ethics and artificial intelligence: suicide prevention on facebook.
\newblock \emph{Philosophy \& Technology}, 31:669--684.

\bibitem[{Goyal et~al.(2021)Goyal, Gao, Chaudhary, Chen, Wenzek, Ju, Krishnan, Ranzato, Guzm\'{a}n, and Fan}]{flores101}
Naman Goyal, Cynthia Gao, Vishrav Chaudhary, Peng-Jen Chen, Guillaume Wenzek, Da~Ju, Sanjana Krishnan, Marc'Aurelio Ranzato, Francisco Guzm\'{a}n, and Angela Fan. 2021.
\newblock The flores-101 evaluation benchmark for low-resource and multilingual machine translation.

\bibitem[{Graham et~al.(2017)Graham, Baldwin, Moffat, and Zobel}]{graham2017can}
Yvette Graham, Timothy Baldwin, Alistair Moffat, and Justin Zobel. 2017.
\newblock Can machine translation systems be evaluated by the crowd alone.
\newblock \emph{Natural Language Engineering}, 23(1):3--30.

\bibitem[{Hassib et~al.(2022)Hassib, Hossam, Sameh, and Torki}]{hassib2022aradepsu}
Mariam Hassib, Nancy Hossam, Jolie Sameh, and Marwan Torki. 2022.
\newblock Aradepsu: Detecting depression and suicidal ideation in arabic tweets using transformers.
\newblock In \emph{Proceedings of the The Seventh Arabic Natural Language Processing Workshop (WANLP)}, pages 302--311.

\bibitem[{Huang et~al.(2014)Huang, Zhang, Chiu, Liu, Li, and Zhu}]{huang2014detecting}
Xiaolei Huang, Lei Zhang, David Chiu, Tianli Liu, Xin Li, and Tingshao Zhu. 2014.
\newblock Detecting suicidal ideation in chinese microblogs with psychological lexicons.
\newblock In \emph{2014 IEEE 11th Intl Conf on Ubiquitous Intelligence and Computing and 2014 IEEE 11th Intl Conf on Autonomic and Trusted Computing and 2014 IEEE 14th Intl Conf on Scalable Computing and Communications and Its Associated Workshops}, pages 844--849. IEEE.

\bibitem[{Jebb et~al.(2021)Jebb, Ng, and Tay}]{jebb2021review}
Andrew~T Jebb, Vincent Ng, and Louis Tay. 2021.
\newblock A review of key likert scale development advances: 1995--2019.
\newblock \emph{Frontiers in psychology}, 12:637547.

\bibitem[{Jha et~al.(2023)Jha, Chan, Orji et~al.}]{jha2023identification}
Smriti Jha, Gerry Chan, Rita Orji, et~al. 2023.
\newblock Identification of risk factors for suicide and insights for developing suicide prevention technologies: A systematic review and meta-analysis.
\newblock \emph{Human Behavior and Emerging Technologies}, 2023.

\bibitem[{Ji et~al.(2022)Ji, Li, Huang, and Cambria}]{ji2022suicidal}
Shaoxiong Ji, Xue Li, Zi~Huang, and Erik Cambria. 2022.
\newblock Suicidal ideation and mental disorder detection with attentive relation networks.
\newblock \emph{Neural Computing and Applications}, 34(13):10309--10319.

\bibitem[{Ji et~al.(2020)Ji, Pan, Li, Cambria, Long, and Huang}]{ji2020suicidal}
Shaoxiong Ji, Shirui Pan, Xue Li, Erik Cambria, Guodong Long, and Zi~Huang. 2020.
\newblock Suicidal ideation detection: A review of machine learning methods and applications.
\newblock \emph{IEEE Transactions on Computational Social Systems}, 8(1):214--226.

\bibitem[{Kim et~al.(2023)Kim, Gwak, Kim, and Gang}]{kim2023identifying}
Junglyun Kim, DongHyeon Gwak, Seonhee Kim, and Moonhee Gang. 2023.
\newblock Identifying the suicidal ideation risk group among older adults in rural areas: Developing a predictive model using machine learning methods.
\newblock \emph{Journal of Advanced Nursing}, 79(2):641--651.

\bibitem[{Kirtley et~al.(2022)Kirtley, van Mens, Hoogendoorn, Kapur, and de~Beurs}]{kirtley2022translating}
Olivia~J Kirtley, Kasper van Mens, Mark Hoogendoorn, Navneet Kapur, and Derek de~Beurs. 2022.
\newblock Translating promise into practice: a review of machine learning in suicide research and prevention.
\newblock \emph{The Lancet Psychiatry}, 9(3):243--252.

\bibitem[{Kwet(2019)}]{kwet2019digital}
Michael Kwet. 2019.
\newblock Digital colonialism: Us empire and the new imperialism in the global south.
\newblock \emph{Race \& Class}, 60(4):3--26.

\bibitem[{Larsen et~al.(2015)Larsen, Cummins, Boonstra, O'Dea, Tighe, Nicholas, Shand, Epps, and Christensen}]{larsen2015use}
Mark~E Larsen, Nicholas Cummins, Tjeerd~W Boonstra, Bridianne O'Dea, Joe Tighe, Jennifer Nicholas, Fiona Shand, Julien Epps, and Helen Christensen. 2015.
\newblock The use of technology in suicide prevention.
\newblock In \emph{2015 37th annual international conference of the IEEE engineering in Medicine and biology society (EMBC)}, pages 7316--7319. IEEE.

\bibitem[{Lee et~al.(2022)Lee, Kang, Kim, and Han}]{lee2022detecting}
Daeun Lee, Migyeong Kang, Minji Kim, and Jinyoung Han. 2022.
\newblock Detecting suicidality with a contextual graph neural network.
\newblock In \emph{Proceedings of the eighth workshop on computational linguistics and clinical psychology}, pages 116--125.

\bibitem[{Lee et~al.(2020)Lee, Park, Kang, Choi, and Han}]{lee2020cross}
Daeun Lee, Soyoung Park, Jiwon Kang, Daejin Choi, and Jinyoung Han. 2020.
\newblock Cross-lingual suicidal-oriented word embedding toward suicide prevention.
\newblock In \emph{Findings of the Association for Computational Linguistics: EMNLP 2020}, pages 2208--2217.

\bibitem[{Lewis et~al.(2011)Lewis, Munro, and Vogel}]{lewis2011crisis}
William Lewis, Robert Munro, and Stephan Vogel. 2011.
\newblock Crisis mt: Developing a cookbook for mt in crisis situations.
\newblock In \emph{Proceedings of the Sixth Workshop on Statistical Machine Translation}, pages 501--511.

\bibitem[{Little et~al.(2016)Little, Roche, Chow, Schenck, and Byam}]{little2016national}
Todd~D Little, Kathleen~M Roche, Sy-Miin Chow, Anna~P Schenck, and Leslie-Ann Byam. 2016.
\newblock National institutes of health pathways to prevention workshop: Advancing research to prevent youth suicide.
\newblock \emph{Annals of internal medicine}, 165(11):795--799.

\bibitem[{Lv et~al.(2015)Lv, Li, Liu, and Zhu}]{lv2015creating}
Meizhen Lv, Ang Li, Tianli Liu, and Tingshao Zhu. 2015.
\newblock Creating a chinese suicide dictionary for identifying suicide risk on social media.
\newblock \emph{PeerJ}, 3:e1455.

\bibitem[{Martinengo et~al.(2019)Martinengo, Van~Galen, Lum, Kowalski, Subramaniam, and Car}]{martinengo2019suicide}
Laura Martinengo, Louise Van~Galen, Elaine Lum, Martin Kowalski, Mythily Subramaniam, and Josip Car. 2019.
\newblock Suicide prevention and depression apps’ suicide risk assessment and management: a systematic assessment of adherence to clinical guidelines.
\newblock \emph{BMC medicine}, 17(1):1--12.

\bibitem[{McCoy et~al.(2016)McCoy, Castro, Roberson, Snapper, and Perlis}]{mccoy2016improving}
Thomas~H McCoy, Victor~M Castro, Ashlee~M Roberson, Leslie~A Snapper, and Roy~H Perlis. 2016.
\newblock Improving prediction of suicide and accidental death after discharge from general hospitals with natural language processing.
\newblock \emph{JAMA psychiatry}, 73(10):1064--1071.

\bibitem[{Mishara and Weisstub(2016)}]{mishara2016legal}
Brian~L Mishara and David~N Weisstub. 2016.
\newblock The legal status of suicide: A global review.
\newblock \emph{International journal of law and psychiatry}, 44:54--74.

\bibitem[{Mishra et~al.(2019)Mishra, Sinha, Sawhney, Mahata, Mathur, and Shah}]{mishra2019snap}
Rohan Mishra, Pradyumn~Prakhar Sinha, Ramit Sawhney, Debanjan Mahata, Puneet Mathur, and Rajiv~Ratn Shah. 2019.
\newblock Snap-batnet: Cascading author profiling and social network graphs for suicide ideation detection on social media.
\newblock In \emph{Proceedings of the 2019 conference of the North American Chapter of the association for computational linguistics: student research workshop}, pages 147--156.

\bibitem[{Morrow et~al.(2022)Morrow, Zamora-Resendiz, Beckham, Kimbrel, Oslin, Tamang, Crivelli, Group et~al.}]{morrow2022case}
Destinee Morrow, Rafael Zamora-Resendiz, Jean~C Beckham, Nathan~A Kimbrel, David~W Oslin, Suzanne Tamang, Silvia Crivelli, Million Veteran Program Suicide Exemplar~Work Group, et~al. 2022.
\newblock A case for developing domain-specific vocabularies for extracting suicide factors from healthcare notes.
\newblock \emph{Journal of psychiatric research}, 151:328--338.

\bibitem[{Moslem et~al.(2023)Moslem, Haque, and Way}]{moslem2023adaptive}
Yasmin Moslem, Rejwanul Haque, and Andy Way. 2023.
\newblock Adaptive machine translation with large language models.
\newblock \emph{arXiv preprint arXiv:2301.13294}.

\bibitem[{Moslem et~al.()Moslem, Way, Haque, and Kelleher}]{moslem1domain}
Yasmin Moslem, Andy Way, Rejwanul Haque, and John~D Kelleher.
\newblock Domain-specific text generation for machine translation.
\newblock \emph{Volume 1: MT Research Track}.

\bibitem[{Munzner()}]{munzner2015visualization}
T.~Munzner.

\bibitem[{O'Brien(2017)}]{o2017machine}
Sharon O'Brien. 2017.
\newblock Machine translation and cognition.
\newblock \emph{The handbook of translation and cognition}, pages 311--331.

\bibitem[{O'dea et~al.(2015)O'dea, Wan, Batterham, Calear, Paris, and Christensen}]{o2015detecting}
Bridianne O'dea, Stephen Wan, Philip~J Batterham, Alison~L Calear, Cecile Paris, and Helen Christensen. 2015.
\newblock Detecting suicidality on twitter.
\newblock \emph{Internet Interventions}, 2(2):183--188.

\bibitem[{of~Health et~al.(2016)}]{national2016policy}
National~Institutes of~Health et~al. 2016.
\newblock Policy on good clinical practice training for nih awardees involved in nih-funded clinical trials.
\newblock \emph{NOT-OD-16-1482017}.

\bibitem[{Okhapkina et~al.(2017)Okhapkina, Okhapkin, and Kazarin}]{okhapkina2017adaptation}
Elena Okhapkina, Valentin Okhapkin, and Oleg Kazarin. 2017.
\newblock Adaptation of information retrieval methods for identifying of destructive informational influence in social networks.
\newblock In \emph{2017 31st International Conference on Advanced Information Networking and Applications Workshops (WAINA)}, pages 87--92. IEEE.

\bibitem[{Organization et~al.(2021)}]{world2021suicide}
World~Health Organization et~al. 2021.
\newblock Suicide worldwide in 2019: global health estimates.

\bibitem[{Orr et~al.(2022)Orr, Van~Kessel, and Parry}]{orr2022ethical}
Martin Orr, Kirsten Van~Kessel, and David Parry. 2022.
\newblock The ethical role of computational linguistics in digital psychological formulation and suicide prevention.
\newblock In \emph{Proceedings of the Eighth Workshop on Computational Linguistics and Clinical Psychology}.

\bibitem[{Ortega and Church(2023)}]{ortega2023research}
John~E. Ortega and Kenneth~W. Church. 2023.
\newblock A research-based guide for the creation and deployment of a low-resource machine translation system.
\newblock In \emph{Proceedings of the International Conference on Recent Advances in Natural Language Processing (RANLP 2023)}, pages 809--819.

\bibitem[{Ott et~al.(2019)Ott, Edunov, Baevski, Fan, Gross, Ng, Grangier, and Auli}]{ott2019fairseq}
Myle Ott, Sergey Edunov, Alexei Baevski, Angela Fan, Sam Gross, Nathan Ng, David Grangier, and Michael Auli. 2019.
\newblock fairseq: A fast, extensible toolkit for sequence modeling.
\newblock \emph{arXiv preprint arXiv:1904.01038}.

\bibitem[{Ouyang et~al.(2022)Ouyang, Wu, Jiang, Almeida, Wainwright, Mishkin, Zhang, Agarwal, Slama, Ray et~al.}]{ouyang2022training}
Long Ouyang, Jeffrey Wu, Xu~Jiang, Diogo Almeida, Carroll Wainwright, Pamela Mishkin, Chong Zhang, Sandhini Agarwal, Katarina Slama, Alex Ray, et~al. 2022.
\newblock Training language models to follow instructions with human feedback.
\newblock \emph{Advances in Neural Information Processing Systems}, 35:27730--27744.

\bibitem[{Putnam(1981)}]{putnam1981reason}
Hilary Putnam. 1981.
\newblock \emph{Reason, truth and history}, volume~3.
\newblock Cambridge University Press.

\bibitem[{Ram{\'\i}rez-Cifuentes et~al.(2020)Ram{\'\i}rez-Cifuentes, Freire, Baeza-Yates, Punt{\'\i}, Medina-Bravo, Velazquez, Gonfaus, and Gonz{\`a}lez}]{ramirez2020detection}
Diana Ram{\'\i}rez-Cifuentes, Ana Freire, Ricardo Baeza-Yates, Joaquim Punt{\'\i}, Pilar Medina-Bravo, Diego~Alejandro Velazquez, Josep~Maria Gonfaus, and Jordi Gonz{\`a}lez. 2020.
\newblock Detection of suicidal ideation on social media: multimodal, relational, and behavioral analysis.
\newblock \emph{Journal of medical internet research}, 22(7):e17758.

\bibitem[{Russell(1905)}]{russell1905denoting}
Bertrand Russell. 1905.
\newblock On denoting.
\newblock \emph{Mind}, 14(56):479--493.

\bibitem[{Sarsam et~al.(2021)Sarsam, Al-Samarraie, Alzahrani, Alnumay, and Smith}]{sarsam2021lexicon}
Samer~Muthana Sarsam, Hosam Al-Samarraie, Ahmed~Ibrahim Alzahrani, Waleed Alnumay, and Andrew~Paul Smith. 2021.
\newblock A lexicon-based approach to detecting suicide-related messages on twitter.
\newblock \emph{Biomedical Signal Processing and Control}, 65:102355.

\bibitem[{Sawhney et~al.(2021)Sawhney, Joshi, Gandhi, and Shah}]{sawhney2021towards}
Ramit Sawhney, Harshit Joshi, Saumya Gandhi, and Rajiv~Ratn Shah. 2021.
\newblock Towards ordinal suicide ideation detection on social media.
\newblock In \emph{Proceedings of the 14th ACM International Conference on Web Search and Data Mining}, pages 22--30.

\bibitem[{Sawhney et~al.(2018)Sawhney, Manchanda, Singh, and Aggarwal}]{sawhney2018computational}
Ramit Sawhney, Prachi Manchanda, Raj Singh, and Swati Aggarwal. 2018.
\newblock A computational approach to feature extraction for identification of suicidal ideation in tweets.
\newblock In \emph{Proceedings of ACL 2018, Student Research Workshop}, pages 91--98.

\bibitem[{Schoene et~al.(2023)Schoene, Ortega, Amir, and Church}]{schoene2023example}
Annika~Marie Schoene, John Ortega, Silvio Amir, and Kenneth Church. 2023.
\newblock An example of (too much) hyper-parameter tuning in suicide ideation detection.
\newblock In \emph{Proceedings of the International AAAI Conference on Web and Social Media}, volume~17, pages 1158--1162.

\bibitem[{Sinha et~al.(2019)Sinha, Mishra, Sawhney, Mahata, Shah, and Liu}]{sinha2019suicidal}
Pradyumna~Prakhar Sinha, Rohan Mishra, Ramit Sawhney, Debanjan Mahata, Rajiv~Ratn Shah, and Huan Liu. 2019.
\newblock \# suicidal-a multipronged approach to identify and explore suicidal ideation in twitter.
\newblock In \emph{Proceedings of the 28th ACM international conference on information and knowledge management}, pages 941--950.

\bibitem[{Snowdon and Choi(2020)}]{snowdon2020undercounting}
John Snowdon and Namkee~G Choi. 2020.
\newblock Undercounting of suicides: where suicide data lie hidden.
\newblock \emph{Global public health}, 15(12):1894--1901.

\bibitem[{Speaks(2010)}]{speaks2010theories}
Jeff Speaks. 2010.
\newblock Theories of meaning.

\bibitem[{Tangherlini(2013)}]{tangherlini2013danish}
Timothy~R Tangherlini. 2013.
\newblock \emph{Danish folktales, legends, and other stories}.
\newblock University of Washington Press.

\bibitem[{Team et~al.(2022{\natexlab{a}})Team, Costa-juss{\`a}, Cross, {\c{C}}elebi, Elbayad, Heafield, Heffernan, Kalbassi, Lam, Licht et~al.}]{team2022no}
NLLB Team, Marta~R Costa-juss{\`a}, James Cross, Onur {\c{C}}elebi, Maha Elbayad, Kenneth Heafield, Kevin Heffernan, Elahe Kalbassi, Janice Lam, Daniel Licht, et~al. 2022{\natexlab{a}}.
\newblock No language left behind: Scaling human-centered machine translation.
\newblock \emph{l{\'\i}nea]. Disponible en: https://github. com/facebookresearch/fairseq/tree/nllb}.

\bibitem[{Team et~al.(2022{\natexlab{b}})Team, Costa-jussà, Cross, Çelebi, Elbayad, Heafield, Heffernan, Kalbassi, Lam, Licht, Maillard, Sun, Wang, Wenzek, Youngblood, Akula, Barrault, Gonzalez, Hansanti, Hoffman, Jarrett, Sadagopan, Rowe, Spruit, Tran, Andrews, Ayan, Bhosale, Edunov, Fan, Gao, Goswami, Guzmán, Koehn, Mourachko, Ropers, Saleem, Schwenk, and Wang}]{nllb2022}
NLLB Team, Marta~R. Costa-jussà, James Cross, Onur Çelebi, Maha Elbayad, Kenneth Heafield, Kevin Heffernan, Elahe Kalbassi, Janice Lam, Daniel Licht, Jean Maillard, Anna Sun, Skyler Wang, Guillaume Wenzek, Al~Youngblood, Bapi Akula, Loic Barrault, Gabriel~Mejia Gonzalez, Prangthip Hansanti, John Hoffman, Semarley Jarrett, Kaushik~Ram Sadagopan, Dirk Rowe, Shannon Spruit, Chau Tran, Pierre Andrews, Necip~Fazil Ayan, Shruti Bhosale, Sergey Edunov, Angela Fan, Cynthia Gao, Vedanuj Goswami, Francisco Guzmán, Philipp Koehn, Alexandre Mourachko, Christophe Ropers, Safiyyah Saleem, Holger Schwenk, and Jeff Wang. 2022{\natexlab{b}}.
\newblock No language left behind: Scaling human-centered machine translation.

\bibitem[{Valeriano et~al.(2020)Valeriano, Condori-Larico, and Sulla-Torres}]{Valeriano2020}
Kid Valeriano, Alexia Condori-Larico, and Josè Sulla-Torres. 2020.
\newblock \href {https://doi.org/10.14569/IJACSA.2020.0110489} {Detection of suicidal intent in spanish language social networks using machine learning}.
\newblock \emph{International Journal of Advanced Computer Science and Applications}, 11(4).

\bibitem[{Vaswani et~al.(2017)Vaswani, Shazeer, Parmar, Uszkoreit, Jones, Gomez, Kaiser, and Polosukhin}]{vaswani2017attention}
Ashish Vaswani, Noam Shazeer, Niki Parmar, Jakob Uszkoreit, Llion Jones, Aidan~N Gomez, {\L}ukasz Kaiser, and Illia Polosukhin. 2017.
\newblock Attention is all you need.
\newblock \emph{Advances in neural information processing systems}, 30.

\bibitem[{Wittgenstein(2010)}]{wittgenstein2010philosophical}
Ludwig Wittgenstein. 2010.
\newblock \emph{Philosophical investigations}.
\newblock John Wiley \& Sons.

\bibitem[{Zalta et~al.(1995)Zalta, Nodelman, Allen, and Perry}]{zalta1995stanford}
Edward~N Zalta, Uri Nodelman, Colin Allen, and John Perry. 1995.
\newblock Stanford encyclopedia of philosophy.

\end{thebibliography}

\appendix

\section{Appendix}
In Table \ref{tab:languages-finetune} we show all available languages for translation based on a list of languages from the Flores dataset \cite{flores101} used for the pre-trained language model.

\begin{supertabular}{lcc}
\label{tab:languages-finetune}
%\caption{\label{tab:languages-finetune} Extended overview of no `language left behind' to show if a dictionary was fine-tuned on a general or domain specific dataset or not at all.}
%\begin{center}
%\begin{tabular}{ l | c |c | c| c}
\textbf{Language} & \textbf{Script} & \textbf{Res.}  \\ 
\hline
Acehnese & Arabic & low  \\
Acehnese & Latin & low  \\
Mesopotamian Arabic & low &  Arabic  \\
% Ta/ʽizzi-Adeni Arabic & Arabic & low & no & no \\
Tunisian Arabic & Arabic & low  \\
Afrikaans & Latin & high  \\
South Levantine Arabic & Arabic & low \\
Akan & Latin  & low  \\
Amharic & Ge`ez & low  \\
North Levantine Arabic & Arabic & low  \\
Modern Standard Arabic & Arabic & high \\
Modern Standard Arabic & Latin & low  \\
Najdi Arabic & Arabic & low  \\
Moroccan Arabic & Arabic & low  \\
Egyptian Arabic & Arabic & low \\
Assamese & Bengali & low  \\
Asturian & Latin & low \\
Awadhi & Devanagari & low  \\
Central Aymara &  Latin & low  \\
South Azerbaijani & Arabic & low  \\
North Azerbaijani &  Latin & low  \\
Bashkir & Cyrillic  & low \\
Bambara & Latin & low \\
Balinese & Latin & low \\
Belarusian & Cyrillic  & low  \\
Bemba & Latin & low \\
Bengali & Bengali & high  \\
Bhojpuri & Devanagari & low \\
Banjar & Arabic & low  \\
Banjar & Latin & low  \\
Standard Tibetan & Tibetan & low \\
Bosnian & Latin & high  \\
Buginese & Latin & low \\
Bulgarian & Cyrillic  & high  \\
Catalan & Latin & high  \\
Cebuano & Latin & low  \\
Czech & Latin & high  \\
Chokew & Latin & low \\
Central Kurdish & Arabic & low  \\
Crimean Tatar & Latin & low\\
Welsh & Latin & low  \\
Danish & Latin & high  \\
German & Latin & high  \\
Southwestern Dinka & Latin & low  \\
Dyula & Latin & low  \\
Dzongkha & Tibetan & low \\
Greek & Greek & high  \\
English & Latin & high \\
Esperanto & Latin & low  \\
Estonian & Latin & high  \\
Basque & Latin & high\\
Ewe & Latin & low  \\
Faroese & Latin & low  \\
Fijian & Latin & low \\
Finnish & Latin & high \\
Fon & Latin & low  \\
French & Latin & high  \\
Friulian & Latin & low \\
Nigerian Fulfulde & Latin & low  \\
Scottish Gaelic & Latin & low  \\
Irish & Latin & low  \\
Galician & Latin & low  \\
Guarani & Latin & low \\
Gujarati & Gujarati & low \\
Haitian Creole & Latin & low  \\
Hausa & Latin & low  \\
Hebrew & Hebrew & low \\
Hindi & Devanagari & high  \\
Chhattisgarhi & Devanagari & low  \\
Croatian & Latin & high  \\
Hungarian & Latin & high \\
Armenian & Armenian & low \\
Igbo & Latin & low  \\
Ilocano & Latin & low  \\
Indonesian & Latin & high  \\
Icelandic Latin & high  \\
Italian & Latin & high  \\
Javanese & Latin & low \\
Japanese & Japanese & high \\
Kabyle & Latin & low  \\
Jingpho & Latin & low  \\
Kamba & Latin & low  \\
Kannada & Kannada & low  \\
Kashmiri & Arabic & low  \\
Kashmiri & Devanagari & low  \\
Georgian  & Georgian & low  \\
Central Kanuri &  Arabic & low  \\
Central Kanuri & Latin & low  \\
Kazakh & Cyrillic  & high  \\
Kabiyè &  Latin & low  \\
Kabuverdianu &  Latin & low  \\
Khmer & Khmer & low  \\
Kikuyu &  Latin & low  \\
Kinyarwanda &  Latin & low  \\
Kyrgyz &  Cyrillic &  low  \\
Kimbundu & Latin & low  \\
Northern Kurdish &  Latin & low  \\
Kikongo & Latin & low  \\
Korean & Hangul & high \\
Lao & Lao & low  \\
Ligurian & Latin & low  \\
Limburgish & Latin & low  \\
Lingala & Latin & low  \\
Lithuanian & Latin & high  \\
Lombard & Latin & low  \\
Latgalian & Latin & low  \\
Luxembourgish & Latin & low  \\
Luba-Kasai & Latin & low  \\
Ganda & Latin & low \\
Luo & Latin & low \\
Mizo & Latin & low  \\
Standard Latvian & Latin & high  \\
Magahi &  Devanagari & low \\
Maithili & Devanagari & low  \\
Malayalam & Malayalam & low  \\
Marathi & Devanagari & low  \\
Minangkabau & Arabic & low  \\
Minangkabau & Latin & low  \\
Macedonian & Cyrillic & high  \\
Plateau Malagasy & Latin & low  \\
Maltese & Latin & high  \\
Meitei & Bengali & low  \\
Halh Mongolian & Cyrillic & low  \\
Mossi & Latin & low  \\
Maori & Latin & low  \\
Burmese & Myanmar & low  \\
Dutch & Latin & high  \\
Norwegian Nynorsk & Latin & low  \\
Norwegian Bokmål  & Latin & low  \\
Nepali & Devanagari & low  \\
Northern Sotho &  Latin & low  \\
Nuer &  Latin & low  \\
Nyanja & Latin & low  \\
Occitan & Latin & low  \\
West Central Oromo & Latin & low  \\
Odia & Oriya  & low \\
Pangasinan & Latin & low  \\
Eastern Panjabi & Gurmukhi & low  \\
Papiamento & Latin & low  \\
Western Persian & Arabic & high\\
Polish &  Latin & high  \\
Portuguese &  Latin & high  \\
Dari & Arabic & low \\
Southern Pashto & Arabic & low  \\
Ayacucho Quechua & Latin & low \\
Romanian & Latin & high  \\
Rundi &  Latin & low  \\
Russian & Cyrillic & high \\
Sango & Latin & low  \\
Sanskrit & Devanagari & low  \\
Santali & Ol Chiki & low  \\
Sicilian & Latin & low  \\
Shan & Myanmar & low \\
Sinhala & Sinhala & low \\
Slovak & Latin & high  \\
Slovenian & Latin & high  \\
Samoan & Latin & low  \\
Shona & Latin & low  \\
Sindhi & Arabic & low  \\
Somali & Latin & low  \\
Southern Sotho & Latin & high  \\
Spanish & Latin & high \\
Tosk Albanian & Latin & high  \\
Sardinian & Latin & low  \\
Serbian & Cyrillic & low  \\
Swati & Latin & low \\
Sundanese & Latin & low  \\
Swedish &  Latin & high \\
Swahili  &  Latin & high  \\
Silesian & Latin & low  \\
Tamil & Tamil & low  \\
Tatar & Cyrillic & low  \\
Telugu & Telugu & low  \\
Tajik & Cyrillic & low  \\
Tagalog &  Latin & high  \\
Thai & Thai & high  \\
Tigrinya & Ge`ez & low \\
Tamasheq & Latin & low \\
Tamasheq & Tifinagh & low  \\
Tok Pisin & Latin & low \\
Tswana & Latin & high  \\
Tsonga & Latin & low \\
Turkmen& Latin & low  \\
Tumbuka & Latin & low  \\
Turkish & Latin & high  \\
Twi  & Latin & low  \\
Central Atlas Tamazight & Tifinagh & low  \\
Uyghur & Arabic & low  \\
Ukrainian & Cyrillic & high  \\
Umbundu &  Latin & high \\
Urdu & Arabic & low  \\
Northern Uzbek &  Latin & high  \\
Venetian & Latin & low \\
Vietnamese & Latin & high  \\
Waray & Latin & low  \\
Wolof & Latin & low  \\
Xhosa & Latin & high \\
Eastern Yiddish & Hebrew & low  \\
Yoruba & Latin & low  \\
Yue Chinese & Han (Traditional) & low  \\
Chinese & Han (Simplified) & high  \\
Chinese & Han (Traditional) & high \\
Standard Malay & Latin & high \\
Zulu & Latin & high  \\
\hline 
%\end{tabular}
%\end{center}
%\vspace*{-0.3cm}
%\caption{List of languages from the Flores dataset \cite{flores101}. used for the pre-trained language model.}
\end{supertabular}

\end{document}